\documentclass[letterpaper, 10 pt, conference]{ieeeconf}  

\IEEEoverridecommandlockouts                              

\usepackage{graphics} 
\usepackage{epsfig} 
\usepackage{mathptmx} 
\usepackage{times} 
\usepackage{amssymb}  
\usepackage{cite}
\usepackage{amsmath,amssymb,amsfonts}
\usepackage{algorithmic}
\usepackage{algorithm}

\title{\LARGE \bf
Personalized Decision Supports based on Theory of Mind Modeling and Explainable Reinforcement Learning
}

\author{Huao Li$^{1}$, Yao Fan$^{1}$, Keyang Zheng$^{1}$, Michael Lewis$^{1}$, Katia Sycara$^{2}$
\thanks{*This research was supported by DARPA award HR001120C0035 and AFOSR/AFRL award FA9550-18-1-0251.}
\thanks{$^{1}$Huao Li, Yao Fan, Keyang Zheng, and Michael Lewis are with the School of Computing and Information,
        University of Pittsburgh, Pittsburgh, PA 15260, USA
        {\tt\small \{hul52,yao.fan,kez20,cmlewis\}@pitt.edu}}%
\thanks{$^{2}$Katia Sycara is with the Robotics Institute, Carnegie Mellon University,
        Pittsburgh, PA 15213, USA
        {\tt\small katia@cs.cmu.edu}}%
}

\begin{document}

\maketitle
\thispagestyle{empty}
\pagestyle{empty}

\begin{abstract}

In this paper, we propose a novel personalized decision support system that combines Theory of Mind (ToM) modeling and explainable Reinforcement Learning (XRL) to provide effective and interpretable interventions. Our method leverages DRL to provide expert action recommendations while incorporating ToM modeling to understand users' mental states and predict their future actions, enabling appropriate timing for intervention. To explain interventions, we use counterfactual explanations based on RL's feature importance and users' ToM model structure. Our proposed system generates accurate and personalized interventions that are easily interpretable by end-users. We demonstrate the effectiveness of our approach through a series of crowd-sourcing experiments in a simulated team decision-making task, where our system outperforms control baselines in terms of task performance. Our proposed approach is agnostic to task environment and RL model structure, therefore has the potential to be generalized to a wide range of applications.

\end{abstract}


\section{Introduction}

Decision Support Systems (DSS) based on Artificial Intelligence (AI) have found broad applications in various domains such as clinical decision-making~\cite{valente2022interpretability} and intelligent tutoring systems~\cite{huang2022supporting}. Recent advancements in deep learning models have enabled AI to conduct more challenging tasks faster than humans. However, the practical applicability of machine learning techniques in DSS has been limited by two major challenges - personalization and explainability. Traditional machine learning models rely on large datasets to build function approximations for general patterns, which is often unfeasible in domains with limited data availability or heterogeneous data distributions. Particularly, when user data is involved, DSS must account for unique task scenarios and user preferences. Therefore, personalized models that capture individual differences among users may be more appropriate~\cite{mcauley2022}. On the other hand, the black-box nature of many ML models often leads to a lack of trust and insufficient transparency~\cite{antoniadi2021current}. It is easier for human users to trust and rely on a DSS if they can interpret the given interventions~\cite{lewis2021deep}. For critical tasks with a high cost of failure, DSS must be transparent about their confidence levels in a given context so that users can selectively use the interventions. Otherwise, users may either overtrust the system and comply with inappropriate interventions or undertrust the system and fail to take advantage of appropriate interventions~\cite{lee2004trust}. 

\begin{figure}[t]
\centerline{\includegraphics[width=0.95\linewidth]{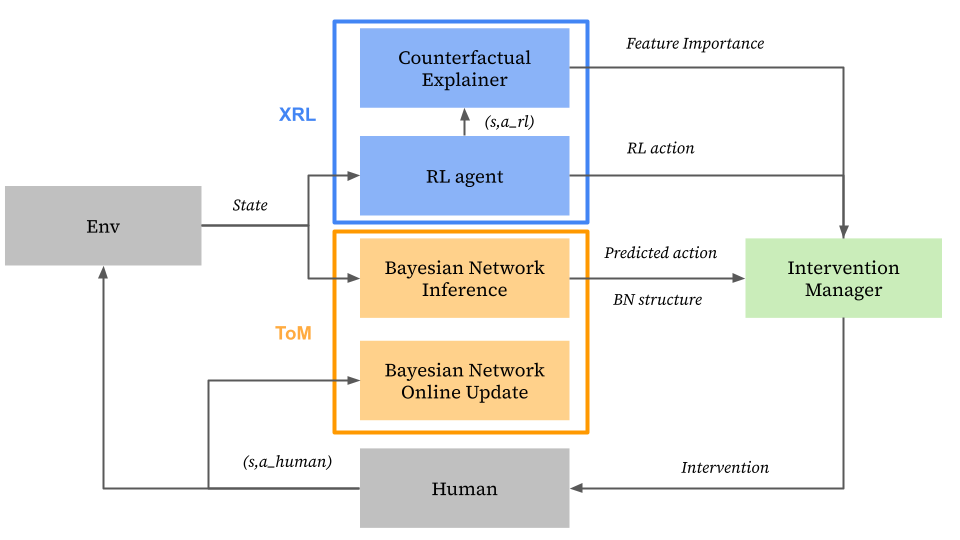}}
\caption{The framework of proposed decision support system.}
\label{framework}
\end{figure}

In this paper, we present a novel decision support system for providing personalized interventions to humans in sequential decision-making tasks to address above mentioned challenges. Our system combines Deep Reinforcement Learning (DRL), Theory of Mind (ToM) modeling and explainable Reinforcement Learning (XRL) methods to provide effective and interpretable interventions, as shown in Fig.~\ref{framework}.
The method is built on a pre-trained DRL agent that provides expert action recommendations, while ToM modeling is used to identify situations where interventions are necessary. To achieve this, we have developed a Bayesian Network-based computational ToM model that takes in periodic observations as evidence and updates online its beliefs about human mental states over the course of interaction. To make our interventions interpretable, we calculate the feature importance of DRL's local decisions by comparing the distance between counterfactual states. We also take into account users' ToM model structure to personalize interventions. 
We evaluate the effectiveness of our system by a series of crowd-sourcing experiments with human participants in a simulated decision-making task. The results showed that our method outperformed various control baselines and significantly improved participants' task performance. 

The contributions of our work are fourfold. First, our method is among the few attempts to extend DRL advice to enhancing human performance. Second, we have evaluated explainable interventions in a crowd-sourcing study, which adds to the body of knowledge in this area. Third, we have shown that incorporating a computational ToM model increases the effectiveness of agent interventions. Fourth, our method is agnostic to task environment and DRL model structure, which makes it easy to generalize to a wide range of applications.

\section{Related Work}

\subsection{Decision Support Systems}


The concept of using computer programs to aid decision-making dates back to earlier works such as Bonini~\cite{bonini1962simulation} and Gorry~\cite{gorry1971framework} in the 1960s and 1970s, respectively. Over time, a significant body of literature has accumulated in this area. More recently, researchers have explored the use of explainable reinforcement learning methods to develop decision support systems, particularly in the context of medical decisions~\cite{antoniadi2021current}. Existing DSS tend to be domain-specific, focusing on specific decisions (e.g. a particular disease as in~\cite{du2022explainable}) or leveraging an existing knowledge base. In contrast, our XRL-based system is designed to be easily generalizable and not limited to a specific domain. Furthermore, our system is tested in the context of sequential decision making, where decision makers are required to make a sequence of decisions with interdependence, setting it apart from existing approaches in the literature.

\subsection{Reinforcement Learning}

Deep Reinforcement Learning has demonstrated remarkable performance in various tasks and domains, including beating the best human GO players~\cite{silver2016mastering}, playing competitive video games~\cite{berner2019dota}, and conversing like humans~\cite{stiennon2020learning}.
Recently, there has been growing interest in using DRL to enhance human performance~\cite{reddy2017accelerating}. Previous research has explored teacher-student interactions, where a teacher agent transfers its learned policies to a student agent to speed up the student's learning~\cite{torrey2013teaching,omidshafiei2019learning}.
Despite various methods being proposed to share knowledge via action advising between artificial agents, little attention has been paid to extending this advising to human learners.
The challenges in making DRL advice useful to humans stem from the different characteristics between human and machine intelligence.
For example, given humans' limited information processing capability, interventions should be given selectively and only when necessary.
While most prior works have used a "budget" to constrain the intervention frequency between agents~\cite{ilhan2019teaching,zhu2020learning}, none have considered building a cognitive model, such as a Theory of Mind model, to identify the appropriate timing and frequency for intervening with humans.

Additionally, humans often require natural language explanations to better understand the interventions, whereas state-action pairs may be sufficient for agent recipients to learn~\cite{williams2013explanation}. A recent survey on XRL~\cite{milani2022survey} proposes a novel taxonomy in which explanations can be generated based on feature importance, learning process, or policy-level methods. Regarding post-hoc explanations for RL agents' local decisions, counterfactual explanations are shown to be valid by revealing what should have been different in an instance to observe a diverse outcome~\cite{guidotti2022counterfactual}.
To address these issues, our proposed DSS only intervenes when expected human action differs from the RL agent's recommendation and provides counterfactual explanations based on the RL agent's feature importance and the structure of the human's ToM model.


\subsection{Computational Theory of Mind Modeling}

Theory of mind refers to the ability to make inferences about the mental states of others, including their desires, beliefs, and intentions~\cite{premack1978does}. This ability is fundamental for human social interactions and everyday activities such as empathy, teamwork, and communication. As intelligent agents increasingly become part of human-agent teams, they will also be expected to possess social intelligence and provide appropriate assistance when needed. To achieve this, AI must maintain an explicit human model capable of estimating hidden mental states from observable behaviors and interaction history. Researchers have developed various methods for building computational Theory of Mind models to enable AI to infer human mental states, including planning-based methods~\cite{breazeal2009embodied,oh2014probabilistic, oh2013prognostic}, Bayesian methods~\cite{baker2009action,baker2017rational,liu2017ten}, and modern machine learning methods such as neural networks~\cite{jain2020predicting,oguntola2021deep,li2023sequential}. However, few attempts have been made to apply these ToM models to enhance the performance of decision support systems. In this paper, we leverage a Bayesian Network (BN)~\cite{pearl1988probabilistic} to estimate human decision-making processes in a simulated search and rescue task and use the model to identify appropriate content and timing for decision support. This constitutes a straightforward yet effective computational model of ToM.

\section{Methods}
\label{method}

\subsection{Explainable Reinforcement Learning}
Reinforcement learning solves sequential decision making problems, which are usually formulated as Markov Decision Processes (MDPs). A MDP is defined by a tuple $\langle S,A,T,R,\gamma \rangle$ where $S$ is a set of environment states, $A$ is a set of available actions, $T(s,a,s') =P(s'|s,a)$ is the transition function between states due to the action $a$, $R: S \to R $ is is a reward function returning a reward for being in state $s$ and performing action $a$, and $\gamma\in[0,1]$ is a discount factor that balances the importance of future rewards. An expert policy $\pi(s,a)=P(a|s)$ can be trained to map states to actions, with the goal of maximizing the expected cumulative discounted reward. Depending on the specific sequential decision making problem, different RL methods can be employed, such as value-based, policy-based, or actor-critic methods. Once an expert policy $\pi$ is obtained, the decision support system (DSS) can provide expert action recommendations $\pi(s)=a_{expert}$ for any given state $s$.

In addition to action recommendations, the DSS must also provide rationale for its behaviors. Using the MDP framework, we define counterfactual explanations as follows: given a policy $\pi$ that selects action $\pi(s) = a$ in state $s$, a counterfactual explanation consists of a state $s\prime$ such that the agent’s policy chooses action $a`$ instead of $a$, i.e., $\pi(s\prime) \neq a$, and such that the difference between $s$ and $s\prime$ is minimal~\cite{guidotti2022counterfactual}. The heuristic search strategy we use to retrieve counterfactual explanations is similar to the local search algorithm used in~\cite{lash2017generalized}. The current state $s$ is perturbed one feature $q$ at a time in order to find the closet counterfactual state $s\prime = s+b e_q$ that leads to a different agent action, where $b$ is the number of perturbations and $e_q$ is a vector that equals one in the $q$-th coordinate and zero elsewhere. The minimal number of perturbations $b_q$ required to yield counterfactual actions is used to quantify the importance or salience of feature $q$ in the local decision making. 


\subsection{Theory of Mind Model}

The Theory of Mind Model we used to estimate the mental states of human participants is based on a Bayesian Network. A Bayesian network $B = (G=(V,E),P)$ is a probabilistic graphical model that represents the conditional dependencies of a set of random variables $X$. The inter-variable dependence structure is represented by nodes $V$ (which refer to the variables) and directed arcs $E$ (which refer to the conditional relationships) in the form of a directed acyclic graph (DAG) $G=(V,E)$. There are two sub-tasks involved in learning a Bayesian network: 1) determining the causal probability relationships as a DAG structure and 2) estimating the parameters of conditional probability distributions. 

Given a set of data $D$ from observed human behaviors, we search for a Bayesian network structure $B_s$ that maximize the probability $P(B_s,D)$. A popular approach to this problem is called score-based learning, which assigns a score to each Bayesian network structure according to a scoring function and finds the structure that optimizes the score~\cite{liu2012empirical}. In this paper, we use a hybrid structure learning method that combines expert knowledge about the task and Hill Climb Search based on Bayesian Dirichlet equivalence (BDeu) score ~\cite{heckerman1995learning}. Once the DAG structure is determined, the maximum likelihood estimator is employed as the parameter learning method.

\renewcommand{\algorithmicrequire}{\textbf{Input:}}
\renewcommand{\algorithmicensure}{\textbf{Output:}}

\begin{algorithm}[hbt!]
\caption{Theory of Mind Model}\label{alg:one}
\begin{algorithmic}[1]
\REQUIRE State observation $s$, Human action $a_h$, Observation window size $w$
\ENSURE Bayesian Network Structure $B_s$, Bayesian Network Parameters $B_{CPD}$, Confidence threshold $threshold$
\STATE Memory $M \gets \O$

\IF{$len(M) > w$}
  \STATE $B_s \gets HillClimb(M)$
  \STATE $threshold \gets UpdateThreshold(M)$
  \STATE $M \gets \O$
    \IF{$B_s changes$}
        \STATE $B_{CPD} \gets MLE(M)$
    \ENDIF
\ELSE{}
\STATE $M.append((s,a_h))$
\STATE $B_{CPD} \gets BayesianEstimator(s,a_h)$ 
\ENDIF

\end{algorithmic}
\end{algorithm}

The algorithm for online updating of the ToM model is presented in Algorithm~\ref{alg:one}. The hyperparameter $w$ represents the length of the moving observation window and is set to the minimum amount of evidence required to initialize the ToM model and estimate belief. As human behavior data is fed into the model, it accumulates the most recent $w$ observations in its memory $M$. Once the number of observations exceeds $w$, the Hill climb search method is used to find the optimal DAG structure with the highest BDeu score given $M$. If the belief model structure is different from the previous estimation, the priors are discarded, and the conditional probability distributions (CPDs) are re-estimated using the maximum likelihood estimator based on the most recent observations in memory. Otherwise, the model parameters are updated online using a Bayesian estimator as more observations come in. Additionally, the confidence threshold is updated by searching for a threshold with the highest prediction accuracy based on $M$.

\subsection{Intervention Manager}

The algorithm for managing and issuing interventions is presented in Algorithm~\ref{alg:two}. When the ToM model is initialized, the Bayesian Network can predict human action $a_h$ with a certain level of confidence, given the current state $s$. If the confidence level is greater than the predefined threshold and the predicted human action is different from the action recommended by the RL model, an intervention will be issued to the human participant. The intervention includes an action recommendation and an explanation for this decision. The action recommendation is the action $a_{expert}$ that the RL agent takes in the current state. The explanation is generated by identifying the most important causal relationship $E$ that is overlooked by human participants, using the feature importance generated by counterfactual explanations and the DAG structure of the ToM model.

\begin{algorithm}[hbt!]
\caption{Intervention Manager}\label{alg:two}
\begin{algorithmic}[1]
\REQUIRE State observation $s$, ToM model $ToM$, Expert RL policy $\pi$
\ENSURE Intervention $I$, Explanation $E$
\STATE $I \gets \O$
\STATE $E \gets \O$

\IF{ToM is initialized}
  \STATE $a_{pred} \gets ToM.inference(s)$
  \STATE $confidence \gets ToM.inference(s)$
  \STATE $a_{expert} \gets \pi(s)$
    \IF{$confidence > threshold \AND a_{pred} \neq a_{expert}$}
        \STATE $I \gets a_{expert}$
        \STATE $E \gets CounterFactual(\pi,s, ToM)$
    \ENDIF
\ENDIF
\STATE $ToM.update(s,a_{human})$

\end{algorithmic}
\end{algorithm}

\section{Experiments}
We implement our proposed method on a simulated sequential decision making task, and evaluate it with human participants in a crowd-sourcing study. 
\subsection{Task Scenario}
The task scenario is based on the urban search and rescue domain, where a team of first responders collaborate to search for and defuse bombs within a limited time frame. The objective of the task is to earn as many points as possible by successfully defusing bombs. At each round, the agent observes three variables: the \textit{type} of the bomb, the \textit{distance} to teammates, and the remaining \textit{time}. Based on these observations, the agent must choose an \textit{action}, either defusing the bomb alone or calling for help from teammates. The reward for the agent is based on the state-action pair, and the episode ends when all 12 bombs have been defused or time has run out. The design of the task requires the agent to consider various state observations and make strategic decisions to maximize performance, as more difficult bombs can only be defused with the help of teammates and calling for help incurs additional time costs proportional to the distance between the agent and their teammates.

\subsection{Experimental Conditions}
To evaluate the effectiveness of our proposed method, we conducted a crowd-sourcing study using the task scenario described above. Participants were randomly assigned to one of the four experimental conditions as follows:
\subsubsection{ToM + XRL}
Interventions are issued based on the ToM and XRL method introduced in Section~\ref{method}.
\subsubsection{XRL only}
Interventions are issued based on XRL output only without considering the mental state of human participants. A random filter is used to control the intervention frequency to the same level as in previous conditions (i.e. only issuing interventions in 9.5\% of the rounds). RL agent's expert action is recommended and the most salient feature is used as the explanation.
\subsubsection{ToM only}
No action recommendation or explanation is provided in this condition. Instead, general tips about ideal strategy of the task scenario are provided based on the human action in last round. The contents of tips are similar to the explanation templates used in previous conditions. 
\subsubsection{No intervention}
No intervention is issued in this condition, which serves as a control baseline.

\subsection{Crowd-sourcing Study}

\begin{figure}[htb]
\centerline{\includegraphics[width=0.95\linewidth]{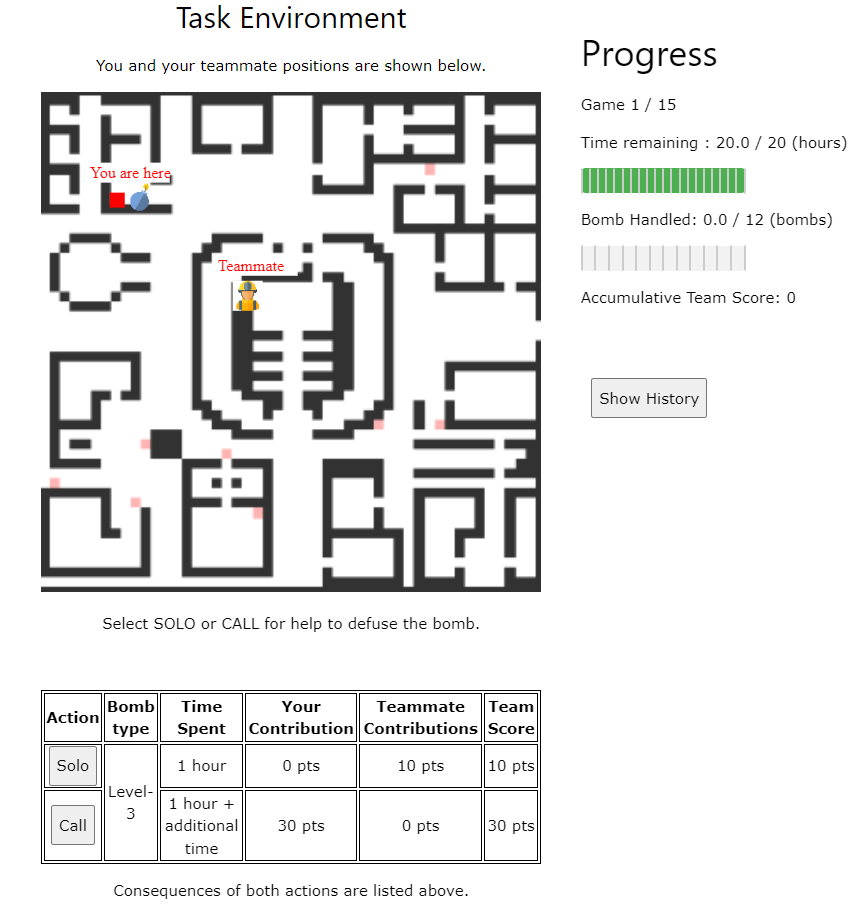}}
\caption{Web interface of the crowd-sourcing study.}
\label{interface}
\end{figure}

In our crowd-sourcing study, participants were required to complete 12 trials (episodes) of the experiment, with the first 3 trials serving as a training session, while the remaining 9 trials were dedicated to actual data collection. To help participants understand the task, detailed instructions were provided beforehand, which included information about the task's background, potential action options, and the detailed payoff table. The interface of the data collection website is shown in Fig.~\ref{interface}. The middle panel displays the human participant's and teammates' locations, which can be used to infer the \textit{distance}. The lower table shows the payoff table for the current round. For instance, if the \textit{bomb type} is level-3, which represents the most difficult bombs, then the score for soloing this bomb would be relatively low (i.e. 10 pts), while the reward for calling for help would be relatively high (i.e. 30 pts). The exact \textit{time} cost of different actions is not displayed to participants and must be learned through experience. The right panel shows the experiment's progress, such as the current trial number, time remaining, bombs handled, and cumulative score for the current trial.

During each round of the game, participants receive information about the \textit{bomb type}, \textit{distance}, and \textit{time} through the interface and must decide on an appropriate \textit{action}. After taking the action, the reward and time cost for that round are displayed. If an intervention is issued, a pop-up window appears as a reminder, with the intervention text displayed above the payoff table. Natural language templates are employed to convert the action recommendations and explanations into English sentences. For instance, if the action recommendation is to \textit{solo} the bomb and the most important feature is \textit{time}, the intervention message suggests the following 'Consider soloing this round. Calling for help may cost too much time and reduce the number of bombs you can attend to.'.

\section{Results}
In total of 247 participants were recruited from Prolific for the crowd-sourcing study. After removing participants who did not complete the task or failed to pass attention checks, we were left with 214 valid data points.

\subsection{Prediction Model}

We began by evaluating the performance of our proposed ToM model in terms of its ability to accurately predict human behaviors and represent human mental states. In this evaluation, we compared our method with classic data-driven machine learning methods as baselines. To exclude the influence of interventions, we only used data from the control group for this comparison, and we divided participants into 10 cross-validation groups to test model generalization. As shown in Table~\ref{pred_table}, our proposed ToM method outperformed the baseline methods in predicting human actions while still preserving adequate explainability. The superior accuracy of our proposed ToM model is mainly due to its personalization to individual participants, taking into account previous observations and updating beliefs and confidence thresholds in real-time. Moreover, the Bayesian Network structure of ToM model is highly interpretable, allowing for effective intervention guidance.

\begin{table}[htbp]
\caption{Performance table of prediction models} 
\begin{center}
\begin{tabular}{ccc}
\hline
Methods    &    Prediction Accuracy          \\ \hline
\textbf{ToM} &\textbf{86.4\%}   \\
Logistic Regression         &75.0\%        \\ 
Decision Tree        &67.3\%        \\ 
Gradient Boosting         &75.5\%       \\ 
Neural Network         &63.6\%        \\  \hline
\end{tabular}

\label{pred_table}
\end{center}
\end{table}

\subsection{Task Performance}

\begin{figure}[htb]
\centerline{\includegraphics[width=0.95\linewidth]{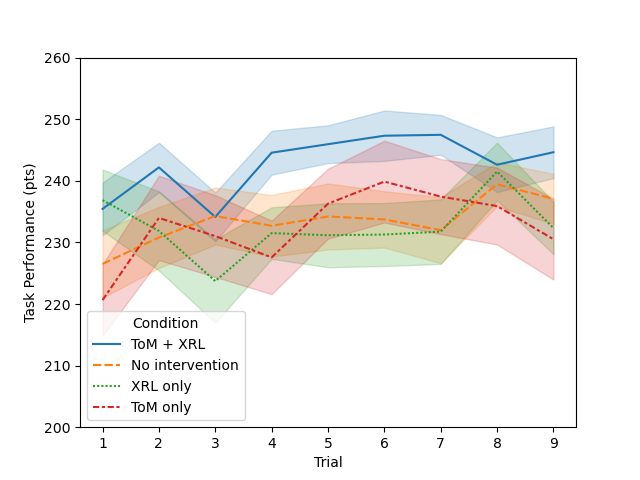}}
\caption{Learning curves of participants' task performance in four experimental conditions. Shaded areas represent standard errors.}
\label{learning_curves}
\end{figure}

Participants' task performance is illustrated in Fig.~\ref{learning_curves}. A mixed ANOVA was conducted, with experimental condition as the between-subject variable and trial as the within-subject (repeated measurements) variable. The results show that participants' task performance increased over time as they gained experience and continued to refine their strategies, with a significant main effect of trial, $F(8,1680) = 2.45, p = .01$. Furthermore, there were significant differences between the experimental conditions, $F(3,210) = 3.34, p = .02$. Post-hoc paired t-tests indicate that the ToM + XRL group exhibited significantly higher task performance compared to the other three groups, with $ps < .05$.

\subsection{Compliance}

We assessed two types of compliance for interventions issued to participants: short-term compliance and long-term compliance. Short-term compliance was determined by whether participants followed the intervention to select solo or call in the current round of the game. This measurement simply quantified whether participants changed their local decisions upon receiving interventions. Long-term compliance, on the other hand, referred to whether participants changed their ToM state after receiving interventions, which quantified the effect of intervention explanations. Specifically, we compared participants' BN model structures before and after a certain intervention to check if the empathized causal relationship appeared. For example, if the intervention says 'Consider soloing this round. Calling for help may cost too much time and reduce the number of bombs you can attend to.', the participant is short-term complied if she indeed chooses to solo in this round, and she is long-term complied if she takes the variable \textit{time} into consideration in consequential trials.

Table~\ref{comp_table} showed that the short-term compliance rate was similar between the ToM + XRL and XRL only groups. The relatively high compliance rate (approximately 90\%) confirmed the effectiveness of intervention presentations. The XRL only group had a slightly higher short-term compliance rate because it may have instructed participants to take actions that they would have taken even without interventions. However, in terms of long-term compliance, the ToM + XRL group performed better than the XRL only condition due to the help of the ToM model. Our proposed method could detect causal relationships overlooked by humans and intervene accordingly, leading to better long-term compliance.

\begin{table}[htbp]
\caption{Compliance rate of issued interventions} 
\begin{center}
\begin{tabular}{ccc}
\hline
Group    &         Short-term Compliance     & Long-term Compliance \\ \hline

ToM + XRL  &86.9\%  & 48.1\%      \\ 
XRL only       &90.1\%  & 22.7\%      \\ \hline
\end{tabular}
\label{comp_table}
\end{center}
\end{table}

\section{Discussions}

Our proposed decision support system provides a novel approach to address the challenge of generating effective and trustworthy interventions for complex decision-making tasks. The combination of Theory of Mind modeling and explainable Reinforcement learning allows the system to provide personalized and adaptive recommendations and explanations of the reasoning behind them. The use of a Bayesian Network to model the mental state of human recipients is a significant advantage, enabling the system to make accurate predictions and provide insights for potential interventions. Furthermore, the system's ability to update the ToM model in real-time based on new observations enables it to provide personalized estimation that evolve as the human mental state changes. Our study also demonstrated the effectiveness of the interventions provided by the proposed system. Specifically, participants who complied with the local optimal action recommendations achieved a better payoff in the short-term. Moreover, the explanations help participants make more rational decisions in the long-term, by highlighting the most salient causal relationships that were overlooked by the human ToM model, resulting in better overall performance. This indicates that our personalized decision support system can be an effective tool for helping individuals make complex decisions in dynamic environments.

However, there are some limitations to our study that need to be addressed in future research. Firstly, the searching methods used for counterfactual states and ToM structures depend on a few assumptions about feature weights and distributions. Future studies could explore advanced searching algorithms, such as optimization-based methods, to remove these assumptions and generalize the method to more complicated scenarios. Secondly, we only evaluated the method in an abstract decision making problem with the Markov property and a discrete action space. Further research can extend the proposed method to real-world application scenarios with more complex and continuous decision-making spaces.

\bibliographystyle{IEEEtran}
\bibliography{IEEEabrv,refs}

\end{document}